\pdfoutput=1

\documentclass[11pt]{article}

\usepackage[]{acl}

\usepackage{times}
\usepackage{latexsym}

\usepackage[T1]{fontenc}

\usepackage[utf8]{inputenc}

\usepackage{microtype}

\usepackage{amsmath,amssymb,amsfonts}
\usepackage{bm}

\usepackage{xcolor,colortbl}
\usepackage{multirow}
\usepackage{booktabs}

\usepackage{graphicx}
\usepackage{subcaption}

\title{Residue-Based Natural Language Adversarial Attack Detection}

\author{Vyas Raina \\
  ALTA Institute, Cambridge University \\
  \texttt{vr313@cam.ac.uk} \\\And
  Mark Gales \\
   ALTA Institute, Cambridge University \\
  \texttt{mjfg@cam.ac.uk} \\}

\begin{document}
\maketitle
\begin{abstract}

Deep learning based systems are susceptible to adversarial attacks, where a small, imperceptible change at the input alters the model prediction. However, to date the majority of the approaches to detect these attacks have been designed for image processing systems. Many popular image adversarial detection approaches are able to identify adversarial examples from embedding feature spaces, whilst in the NLP domain existing state of the art detection approaches solely focus on input text features, without consideration of model embedding spaces. This work examines what differences result when porting these image designed strategies to Natural Language Processing (NLP) tasks - these detectors are found to not port over well. This is expected as NLP systems have a very different form of input: discrete and sequential in nature, rather than the continuous and fixed size inputs for images. As an equivalent model-focused NLP detection approach, this work proposes a simple sentence-embedding "residue" based  detector to identify adversarial examples. On many tasks, it out-performs ported image domain detectors and recent state of the art NLP specific detectors~\footnote{Code is available at: \url{https://github.com/rainavyas/NAACL-2022-Residue-Detector}}.
\end{abstract}

\section{Introduction}

In the last decade deep learning based models have demonstrated success in a wide range of application areas, including Natural Language Processing (NLP)~\citep{DBLP:journals/corr/VaswaniSPUJGKP17} and object recognition~\citep{DBLP:journals/corr/HeZRS15}.  These systems may be deployed in mission critical situations, where there is the requirement for a high level of robustness. However, \citet{42503} demonstrated that deep models have an inherent weakness: small perturbations in the input can yield significant, undesired, changes in the output from the model. 
These input perturbations were termed adversarial examples and their generation adversarial attacks. 

Adversarial attacks have been developed for systems operating in various domains: image systems~\citep{DBLP:journals/corr/abs-2008-04094, DBLP:journals/corr/abs-1712-03141, DBLP:journals/corr/abs-1912-01667} and NLP systems~\citep{DBLP:journals/corr/LinMBHPRDZ14, DBLP:journals/corr/SamantaM17, DBLP:journals/corr/RosenbergSRE17}. The characteristics of the input can be very different between these application domains. Broadly, the nature of inputs can be described using two key attributes: static (fixed length) vs sequential and continuous vs discrete. Under this categorisation, image inputs are continuous and static, whilst NLP inputs are discrete and sequential. This work argues that due to the fundamental differences in the input and resulting  adversarial perturbations in the different domains, adversarial attack behaviour can vary significantly from one domain to another. Hence, the extensive research on exploring and understanding adversarial perturbation behaviour in the continuous, static world of image systems does not necessarily transfer well to the NLP tasks.

For adversarial attack generation, a number of specific NLP attacks have been proposed that are designed for NLP
task inputs~\citep{DBLP:journals/corr/LinMBHPRDZ14, DBLP:journals/corr/SamantaM17, DBLP:journals/corr/RosenbergSRE17, DBLP:journals/corr/abs-1801-02950, DBLP:journals/corr/PapernotMSH16, DBLP:journals/corr/GrossePM0M16, 10.1145/3219819.3219909, DBLP:journals/corr/abs-1803-01128, DBLP:journals/corr/abs-1808-08744, DBLP:journals/corr/abs-1809-01829, Raina2020, DBLP:journals/corr/JiaL17, DBLP:journals/corr/abs-1808-08609, DBLP:journals/corr/abs-1809-02079, ribeiro-etal-2018-semantically, iyyer-etal-2018-adversarial, DBLP:journals/corr/abs-1710-11342}. 
However, there has been less research on developing defence schemes. These defence strategies  can be split into two main groups: \textit{model modification}, where the model or data is  altered  at  training  time (e.g. adversarial training~\citep{yoo2021improving})  and \textit{detection},  where  external  systems or algorithms are applied to trained models to identify adversarial attacks.  As model modification approaches demand re-training of models, detection approaches are usually considered easier for implementation on deployed systems and thus are often preferred. Hence, this work investigates the portability of popular detection approaches designed for image systems to NLP systems. Furthermore, this work introduces a specific NLP detection approach that exploits the discrete nature of the inputs for NLP systems. This approach out-performs standard schemes designed for image adversarial attack detection, as well as other NLP detection schemes.

The proposed NLP specific detection approach will be referred to as \textit{residue detection}, as it is shown that adversarial attacks in the discrete, word sequence  space result in easily detectable residual components in the sentence embedding space. This residue can be easily detected using a simple linear classifier operating in the encoder embedding space. In addition, this work shows that even when an adversary has knowledge of the linear residual detector, they can only construct attacks at a fraction of the original strength. Hence this work argues that realistic (word level, semantically similar) adversarial perturbations at the natural language input of NLP systems leave behind easily detectable residue in the sentence embedding. Interestingly, the residue detection approach is shown to perform poorly when used to detect attacks in the image domain, supporting the hypothesis that the nature of the input has an important influence on the design of effective defence strategies.

\section{Related Work}\label{sec:related}

Previous work in the image domain has analysed the output of specific layers in an attempt to identify adversarial examples or adversarial subspaces. First, \citet{feinman2017detecting} proposed that adversarial subspaces have a lower probability density, motivating the use of the Kernel Density (KD) metric to detect the adversarial examples. Nevertheless, \citet{DBLP:journals/corr/abs-1801-02613} found Local Intrinsic Dimensionality (LID) was a better metric in defining the subspace for more complex data. In contrast to the local subspace focused approaches of KD and LID, \citet{carrara_feat} showed that trajectories of hidden layer features can be used to train a  LSTM network to accurately discriminate between authentic and adversarial examples. Out performing all previous methods, \citet{lee2018simple} introduced an effective detection framework using Mahalanobis Distance Analysis (MDA), where the distance is calculated between a test sample and the closest class-conditional Gaussian distribution in the space defined by the output of the final layer of the classifier. \citet{DBLP:journals/corr/LiL16e} also explored using the output of convolutional layers for image classification systems to identify statistics that distinguish adversarial samples from original samples. They find that by performing a PCA decomposition the statistical variation in the least principal directions is the most significant and can be used to separate original and adversarial samples. However, they argue this is ineffective as an adversary can easily suppress the tail distribution. Hence, \citet{DBLP:journals/corr/LiL16e} extract statistics from the convolutional layer output to train a cascade classifier to separate the original and adversarial samples. Most recently, \citet{mao2019learning} avoid the use of artificially designed metrics and combine the adversarial subspace identification stage and the detecting adversaries stage into a single framework, where a parametric model adaptively learns the deep features for detecting adversaries. 

In contrast to the embedding space detection approaches, \citet{DBLP:journals/corr/abs-1909-06872} shows that influence functions combined with Nearest Neighbour distances perform comparably or better than the above standard detection approaches. Other detection approaches have explored the use of uncertainty: \citet{smith2018understanding} argues that adversarial examples are out of distribution and do not lie on the manifold of \textit{real} data. Hence, a discriminative Bayesian model's epistemic (model) uncertainty should be high. Therefore, calculations of the model uncertainty are thought to be useful in detecting adversarial examples, independent of the domain. However, Bayesian approaches aren't always practical in implementation and thus many different approaches to approximate this uncertainty have been suggested in literature~\citep{Leibig2017LeveragingUI, Gal2016UncertaintyID, gal2016dropout}.

There are a number of existing NLP specific detection approaches. For character level attacks, detection approaches have exploited the grammatical~\citep{DBLP:journals/corr/SakaguchiPD17} and spelling~\citep{MAYS1991517, 10.5555/1699648.1699670} inconsistencies to identify and detect the adversarial samples. However, these character level attacks are unlikely to be employed in practice due to the simplicity with which they can be detected. Therefore, detection approaches for the more difficult \textit{semantically} similar attack samples are of greater interest, where the meaning of the textual input is maintained without compromising the spelling or grammatical integrity. To tackle such word-level, semantically similar examples, \citet{DBLP:journals/corr/abs-1909-03084} designed a discriminator to classify each token representation as part of an adversarial perturbation or not, which is then used to `correct' the perturbation. Other detection approaches~\citep{Raina2020, han2020adversarial, DBLP:journals/corr/abs-1808-08609} have shown some success in using perplexity to identify adversarial textual examples. Most recently, \citet{DBLP:journals/corr/abs-2004-05887} achieved state of the art performance with the Frequency Guided Word Substitution (FGWS) detector, where a change in model prediction after substituting out low frequency words is revealing of adversarial samples. 

\section{Adversarial Attacks} \label{sec:attack}

An adversarial attack is defined as an imperceptible change to the input that causes an undesired change in the output of a system. Often, an attack is found for a specific data point, $\mathbf x$. Consider a classifier $\mathcal F_{\hat\theta}$, with parameters $\hat\theta$, that predicts a class label for an input data point, $\mathbf x$, sampled from the input distribution $\mathcal X$. A successful adversarial attack is where a perturbation $\bm\delta$ at the input causes the system to miss-classify,
\begin{equation} \label{eqn:indv}
    \mathcal F_{\hat{\theta}}(\mathbf x+\bm\delta)\neq \mathcal F_{\hat{\theta}}(\mathbf x).
\end{equation}

When defining adversarial attacks, it is important consider the interpretation of an \textit{ imperceptible} change. Adversarial perturbations are not considered effective if they are easy to detect. Hence, the size of the perturbation must be constrained:
\begin{equation} \label{eqn: general constraint}
    \mathcal G(\mathbf x, \mathbf x+\bm\delta) \leq \epsilon,
\end{equation}
where the function $\mathcal G()$ describes the form of constraint and $\epsilon$ is a selected threshold of imperceptibility. Typically, when considering continuous space inputs (such as images), a popular form of the constraint of Equation \ref{eqn: general constraint}, is to limit the perturbation in the $l_p$ norm, with $p\in[1,\infty)$, e.g. $||\bm\delta ||_p \leq \epsilon$.

For whitebox attacks in the image domain, the dominant attack approach has proven to be Projected Gradient Descent (PGD) \citep{DBLP:journals/corr/KurakinGB16a}. The PGD approach, iteratively updates the adversarial perturbation, $\bm\delta$, initialised as $\bm\delta_0 = \mathbf 0$. Each iterative step moves the perturbation in the direction that maximises the loss function, $\mathcal L$, used in the training of the model, 
\begin{equation} \label{eqn:pgd}
    \bm\delta_{i+1} = \texttt{clip}_{\epsilon}(\bm\delta_i + \alpha\nabla_{\bm\delta_i} \mathcal L(\mathbf x + \bm\delta_i;\hat\theta)),
\end{equation}
where $\alpha$ is an arbitrary step-size parameter and the \textit{clipping} function, $\texttt{clip}_\epsilon$, ensures the imperceptibility constraint of Equation \ref{eqn: general constraint} is satisfied. 

When considering the NLP domain, a sequential, discrete input of $L$ words, can be explicitly represented as,
\begin{equation} \label{eqn:wordsequence}
    \mathbf x = w_{1:L} = w_1, w_2, \hdots, w_{L-1}, w_L,
\end{equation}
where, the discrete word tokens, $w_{1:L}$, are often mapped to a continuous, sequential word embedding~\citep{devlin2019bert} space,
\begin{equation} \label{eqn:embedding}
    \mathbf h_{1:L} = \mathbf h_1, \mathbf h_2, \hdots, \mathbf  h_{L-1}, \mathbf h_L.
\end{equation}
Attacks must take place in the discrete text space, 
\begin{equation}
    \mathbf x+\bm\delta = w'_{1:L'} = w'_1, w'_2, \hdots, w'_{L'-1}, w_{L'}.
\end{equation}
This requires a change in the interpretation of the perturbation $\bm\delta$. It is not simple to define an appropriate function ${\cal G}()$ in Equation \ref{eqn: general constraint} for word sequences. Perturbations can be measured at a character or word level. Alternatively, the perturbation could be measured in the vectorized embedding space (Equation \ref{eqn:embedding}), using for example $l_p$-norm based \citep{43405} metrics or cosine similarity \citep{10.1007/978-3-030-11012-3_26}, which have been used in the image domain. However, constraints in the embedding space do not necessarily achieve imperceptibility in the original word sequence space. The simplest approach is to use a variant of an \textit{edit-based} measurement~\citep{DBLP:journals/corr/abs-1812-05271}, ${\cal L}_{\tt e}()$, which counts the number of changes between the original sequence, $w_{1:L}$ and the adversarial sequence $w'_{1:L'}$, where a change is a swap/addition/deletion, and ensures it is smaller than a maximum number of changes, $N$,
\begin{equation} \label{eqn:nlp constraint}
    {\cal L}_{\tt e}(w_{1:L}, w'_{1:L'}) \leq N.
\end{equation}

For the NLP adversarial attacks this work only examines word-level attacks, as these are considered more difficult to detect than character-level attacks. As an example, for an input sequence of $L$ words, a $N$-word substitution adversarial attack, $w'_{1:N}$, applied at word positions $n_1, n_2, \hdots, n_{N} $ gives the adversarial output, $w'_{1:L'}$
\begin{align} \label{eqn: subst}
    w'_{1:L'} &= w_1, \hdots, w_{n_1-1}, w'_{1}, w_{n_1+1},  \hdots, \nonumber \\&\hspace{1em} w_{n_N-1}, w'_{N}, w_{n_N+1}, \hdots, w_L. 
\end{align}
The challenge is to select which words to replace, and what to replace them with. A simple yet effective substitution attack approach that ensures a small change in the semantic content of a sentence is to use saliency to rank the word positions, and to use word synonyms for the substitutions~\citep{DBLP:conf/acl/RenDHC19}. This attack is termed Probability Weight Word Saliency (PWWS). The highest ranking word word can be swapped for a synonym from a pre-selected list of given synonyms. The next most highly ranked word is substituted in the same manner and the process is repeated till the required $N$ words have been substituted.

The above approach is limited to attacking specific word sequences and so cannot easily be generalised to universal attacks~\citep{DBLP:journals/corr/Moosavi-Dezfooli16}, where the same perturbation is used for all inputs. For this situation, a simple solution is concatenation~\citep{DBLP:journals/corr/abs-1804-06473, DBLP:journals/corr/abs-1808-08744}, where for example, the same $N$-length sequence of words is appended to each input sequence of words, as described in \citet{Raina2020}. Here,
\begin{equation} \label{eqn: concat}
  w'_{1:L'} =  w_1,  \hdots, w_L, w'_{1}, \hdots, w'_{N}.
\end{equation}

In both the substitution attack (Equation~\ref{eqn: subst}) and the concatenation attack (Equation~\ref{eqn: concat}), the size of the attack can be measured using the number of edits, ${\cal L}_{\tt e}(w_{1:L},w'_{1:L'})=N$.
\section{Adversarial Attack Detection} \label{sec:detection}

For a deployed system, the easiest approach to defend against adversarial attacks is to use a detection process to identify adversarial examples without having to modify the existing system.

For the image domain Section \ref{sec:related} discusses many of the standard detection approaches. In this work, we select two distinct approaches that have been generally successful: uncertainty~\citep{smith2018understanding}, where adversarial samples are thought to result in greater epistemic uncertainty and Mahalanobis Distance~\citep{lee2018simple}, where the Mahalanobis distance in the embedding space is indicative of how out of distribution a sample is (adversarial samples are considered more out of distribution). In the NLP domain, when excluding trivial grammar and spelling based detectors, perplexity based detectors can be used~\citep{Raina2020}. Many other NLP specific detectors~\citep{DBLP:journals/corr/abs-1909-03084, han2020adversarial, DBLP:journals/corr/abs-1808-08609} have been proposed, but \citet{DBLP:journals/corr/abs-2004-05887}'s FGWS detector is considered the state of art and is thus selected for comparison. Here low frequency words in an input are substituted for higher frequency words and the change in model prediction is measured - adversarial samples are found to generally have a greater change. This work introduces a further NLP specific detector: \textit{residue detection}, described in detail in Section \ref{sec:residue}. 

When considering any chosen detection measure $\mathcal  F_{\tt d}$, a threshold $\beta$ can be selected to decide whether an input, $w_{1:L}$, is adversarial or not, where ${\cal F}_{\tt d}(w_{1:L}) > \beta$, implies that $w_{1:L}$ is an adversarial sample. To assess the success of the adversarial attack detection processes, precision-recall curves are used. For the binary classification task of identifying an input as adversarially attacked or not, at a given threshold $\beta$, the precision and recall values can be computed as $\text{prec}=\text{TP}/\text{TP}+\text{FP}$ and $\text{rec}=\text{TP}/\text{TP}+\text{FN}$, where TP, FP and FN are the standard true-positive, false-positive and false-negative definitions. A single point summary of precision-recall curves is given with the F$_1$ score.

\subsection{Residue Detection} \label{sec:residue}

In this work we introduce a new NLP detection approach, \textit{residue detection}, that aims to  exploit the nature of the  NLP input space, discrete and sequential. Here we make two hypotheses:

\begin{enumerate}
    \item Adversarial samples in an encoder embedding space result in larger components (\textit{residue}) in central PCA eigenvector components than original examples.
    \item The residue is only significant (detectable) for systems operating on discrete data (e.g. NLP systems).
\end{enumerate}
\noindent The rationale behind these hypotheses is discussed next.

Deep learning models typically consist of many layers of non-linear activation functions. For example, in the NLP domain systems are usually based on layers of the Transformer architecture~\citep{DBLP:journals/corr/VaswaniSPUJGKP17}. The complete end-to-end model $\mathcal F_{\hat{\theta}}()$ can be treated as a two stage process, with an initial set of layers forming the \textit{encoding} stage, $\mathcal F_{\tt en}()$ and the remaining layers forming the \textit{output} stage, $\mathcal F_{\tt cl}()$, i.e. $F_{\hat{\theta}}(\mathbf x) = \mathcal F_{\tt cl}\left(\mathcal F_{\tt en}(\mathbf x)\right)$.

If the encoding stage of the end-to-end classifier is sufficiently powerful, then the embedding space $\mathcal F_{\tt en}(\mathbf x)$ will have compressed the useful information into very few dimensions, allowing the output stage to easily separate the data points into classes (for classification) or map the data points to a continuous value (for regression). A simple Principal Component Analysis (PCA) decomposition of this embedding space can be used to visualize the level of compression of the useful information. The PCA directions can be found using the eigenvectors of the covariance matrix, $\mathbf C$, of the data in the encoder embedding space. If $\{\mathbf q_i\}_{i=1}^d$, where $d$ is the dimension of the encoder embedding space, represent the eigenvectors of $\mathbf C$ ordered in descending order by the associated eigenvalue in magnitude, then it is expected that almost all \textit{useful} information is contained within the first few principal directions, $\{\mathbf q_i\}_{i=1}^p$, where $p\ll d$. Hence, the output stage, $\mathcal F_{\tt cl}()$ will implicitly use only these \textit{useful} components. The impact of a successful adversarial perturbation, $\mathcal F_{\tt en}(\mathbf x +\bm\delta)$, is the significant change in the components in the principal eigenvector directions $\{\mathbf q_i\}_{i=1}^p$, to allow fooling of the output stage. Due to the complex nature of the encoding stage and the out of distribution nature of the adversarial perturbations, there are likely to be residual components in the non-principal  $\{\mathbf q_i\}_{i=p+1}^d$ eigenvector directions. These perturbations in the non-principal directions are likely to be more significant for the \textit{central} eigenvectors, as the encoding stage is likely to almost entirely compress out components in the least principal eigenvector directions, $\{\mathbf q_i\}_{i=d'+1}^d$, where $d'\approx d$. Hence, $\{\mathbf q_i\}_{i=p+1}^{d'}$ can be viewed as a subspace containing adversarial attack residue that can be used to identify adversarial examples.

The existence of adversarial attack residue in the central PCA eigenvector directions, $\{\mathbf q_i\}_{i=p+1}^{d'}$, suggests that in the encoder embedding space, $\mathcal F_{\tt en}(\mathbf x)$, adversarial and original examples are linearly separable. This motivates the use of a simple linear classifier as an adversarial attack detector,
\begin{equation} \label{eqn: lin detect residue}
    P(\text{adv}|\mathbf x) = \sigma(\mathbf W\mathcal F_{\tt en}(\mathbf x)+ b),
\end{equation}
where $\mathbf W$ and $b$ are the parameters of the linear classifier to be learnt and $\sigma$ is the sigmoid function.

The above argument cannot predict how significant the residue in the central eigenvector space is likely to be. For the discrete space NLP attacks, the input perturbations are semantically small, whilst for continuous space image attacks the perturbations are explicitly small using a standard $l_p$-norm. Hence, it is hypothesised that NLP perturbations cause larger errors to propagate through the system, resulting in more significant residue in the encoder embedding space than that for image attacks. Thus, the residue technique is only likely to be a feasible detection approach for discrete text attacks.

The hypotheses made in this section are analysed and empirically verified in Section \ref{sec:analysis}. 

\section{Experiments}

\subsection{Experimental Setup}

Table \ref{tab:data} describes four NLP classification datasets: IMDB~\citep{maas-EtAl:2011:ACL-HLT2011}; Twitter~\citep{saravia-etal-2018-carer}; AG News~\citep{NIPS2015_250cf8b5} and DBpedia~\citep{NIPS2015_250cf8b5}. Further, a regression dataset, Linguaskill-Business (L-Bus)~\citep{chambers-2011-bulats} is included. The L-Bus data is from a multi-level prompt-response free speaking test i.e. candidates from a range of proficiency levels provide open responses to prompted questions. Based on this audio input a system must predict a score of 0-6 corresponding to the 6 CEFR~\citep{cefr-2001} grades. This audio data was transcribed using an Automatic Speech Recognition system with an average word error rate of 19.5\%.

\begin{table}[h!]
    \centering
    \begin{small}
    \begin{tabular}{lrrr}
    \toprule
        Dataset & \#Train & \#Test & \#Classes
        \\ \midrule
         IMDB& 25,000 & 25,000 & 2\\
         Twitter& 16,000 & 2000 &6\\
         AG News& 120,000 & 7600 & 4\\
         DBpedia & 560,000 & 70,000 &14 \\
         L-Bus & 900 & 202 & 1\\
         \bottomrule
    \end{tabular}
    \end{small}
    \caption{NLP Datasets.}
    \label{tab:data}
\end{table}

All NLP task models were based on the Transformer encoder architecture~\citep{DBLP:journals/corr/VaswaniSPUJGKP17}. Table \ref{tab:perf} indicates the specific architecture used for each task and also summarises the classification and regression performance for the different tasks. For classification tasks, the performance is measured by top 1 accuracy, whilst for the regression task (L-Bus), the performance is measured using Pearson Correlation Coefficient (PCC).

\begin{table}[h!]
    \centering
    \begin{small}
    \begin{tabular}{lrr}
    \toprule
        Dataset & Transformer & Performance \\ \midrule
         IMDB & BERT & Acc: 93.8\% \\
         Twitter & ELECTRA & Acc: 93.3\%\\
         AG News & BERT & Acc: 94.5\%\\
         DBpedia & ELECTRA & Acc: 99.2\%\\
         L-Bus & BERT & PCC: 0.749\\
         \bottomrule
    \end{tabular}
    \end{small}
    \caption{Performance of models (BERT~\citep{DBLP:journals/corr/abs-1810-04805}, ELECTRA~\citep{DBLP:journals/corr/abs-2003-10555}).}
    \label{tab:perf}
\end{table}

Table \ref{tab:attack} shows the impact of realistic adversarial attacks on the tasks: substitution (sub) attack (Equation \ref{eqn: subst}), which replaces the $N$ most salient tokens with a synonym defined by WordNet\footnote{\url{https://wordnet.princeton.edu/}}, as dictated by the PWWS attack algorithm described in Section \ref{sec:attack}; or a targeted universal concatenation (con) attack (Equation \ref{eqn: concat}), used for the regression task on the L-Bus dataset, seeking to maximise the average score output from the system by appending the same $N$ words to the end of each input. For classification tasks, the impact of the adversarial attack is measured using the fooling rate, the fraction of originally correctly classified points, misclassified after the attack, whilst for the regression task, the impact is measured as the average increase in the output score.

\begin{table}[h!]
    \centering
    \begin{small}

    \begin{tabular}{lrrr}
    \toprule
        Dataset & Attack & $N$ & Impact \\ \midrule
         IMDB & sub & 25 & Fool: 0.70\\
         Twitter &sub  & 6& Fool: 0.77\\
         AG News  & sub & 40& Fool: 0.65\\
         DBpedia  & sub& 25 & Fool: 0.52\\
         L-Bus  & con& 3 & Score: $+0.51$\\
         \bottomrule
    \end{tabular}
        
    \end{small}
    \caption{Impact of different $N$-word adversarial attacks.}
    \label{tab:attack}
\end{table}

\subsection{Results}

Section \ref{sec:residue} predicts that adversarial attacks in the discrete text space leave residue in a system's encoder embedding space that can be detected using a simple linear classifier. Hence, using the 12-layer Transformer encoder's output $\texttt{CLS}$ token embedding as the encoder embedding space for each dataset's trained system (Table \ref{tab:perf}), a simple linear classifier, as given in Equation \ref{eqn: lin detect residue}, was trained~\footnote{lr=0.02, epochs=20, batch size=200, \#769 parameters} to detect adversarial examples from the adversarial attacks given for each dataset in Table \ref{tab:attack}. The training of the detection linear classifier was performed on the training data (Table \ref{tab:data}) augmented with an equivalent adversarial example for each original input sample in the dataset. Using the test data samples augmented with adversarial examples (as defined by Table \ref{tab:attack}), Table \ref{tab:detect_perf} compares the efficacy of the linear residue detector to other popular detection strategies~\footnote{Detection Strategies: Mahalanobis Distance (MD) used the same train-test split as the residue approach; Perplexity (Perp) was calculated using the language model from~\citet{7472829}; Uncertainty (Unc) used the best measure out of mutual information, confidence, KL-divergence, expected entropy and entropy of expected and reverse mutual information; and FGWS was implemented using the code given at \url{https://github.com/maximilianmozes/fgws}.} (from Section \ref{sec:detection}) using the best F$_1$ score. It is evident from the high F-scores, that for most NLP tasks the linear detection approach is better than other state of the art NLP specific and ported image detection approaches. 

\begin{table}[h!]
    \centering
\begin{small}

    \begin{tabular}{lrrrrr}
    \toprule
        Dataset & Res & Perp & FGWS & MD & Unc\\ \midrule
         IMDB & \textbf{0.91} &0.68 & 0.87& 0.77 & 0.75 \\
         Twitter & \textbf{0.84} & 0.67& 0.76& 0.77 & 0.78\\
         AG News & \textbf{0.95} & 0.69& 0.89 &0.78 & 0.75\\
         DBpedia & 0.80  & 0.67& 0.82 & 0.78&\textbf{0.90}\\
         L-Bus & \textbf{0.99} & 0.68&0.91 & 0.75 & 0.81\\
         \bottomrule
    \end{tabular}

\end{small}
    \caption{F$_1$-score performance of detection approaches.}
    \label{tab:detect_perf}
\end{table}

However, an adversary may have knowledge of the detection approach and may attempt to design an attack that directly avoids detection. Hence, for each dataset, the attack approaches were repeated with the added constraint that any attack words that resulted in detection were rejected. The impact of attacks that suppress detection have been presented in Table \ref{tab:suppress}. Generally, it is shown across all NLP tasks that an adversary that attempts to avoid detection of its residue by a previously trained linear classifier, can only generate a significantly less powerful adversarial attack.

\begin{table}[h!]
    \centering
    \begin{small}

    \begin{tabular}{lrr}
    \toprule
        Dataset & Without & With \\ \midrule
         IMDB & 0.70 & 0.19\\
         Twitter & 0.77 & 0.23\\
         AG News & 0.65& 0.16\\
         DBpedia &0.52 & 0.14\\ \midrule
         L-Bus & $+0.51$ & $+0.23$\\
         \bottomrule
    \end{tabular}
    
    \end{small}
    \caption{Fooling rate (classification) or score (regression) \textit{with} and \textit{without} attack modified to avoid detection.}
    \label{tab:suppress}
\end{table}

\subsection{Analysis} \label{sec:analysis}

\begin{figure}[h!]
    \centering
    \includegraphics[width=1.0\linewidth]{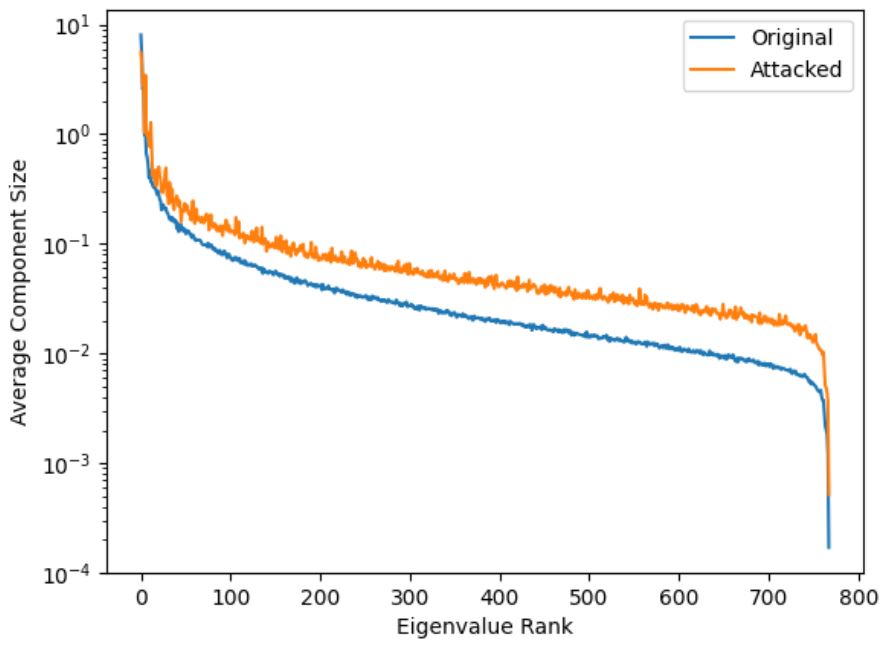}
    \caption{Encoder Embedding Space Residue Plot.}
    \label{fig:nlp decomp}
\end{figure}

\begin{figure}[h!]
    \centering
    \includegraphics[width=1.0\linewidth]{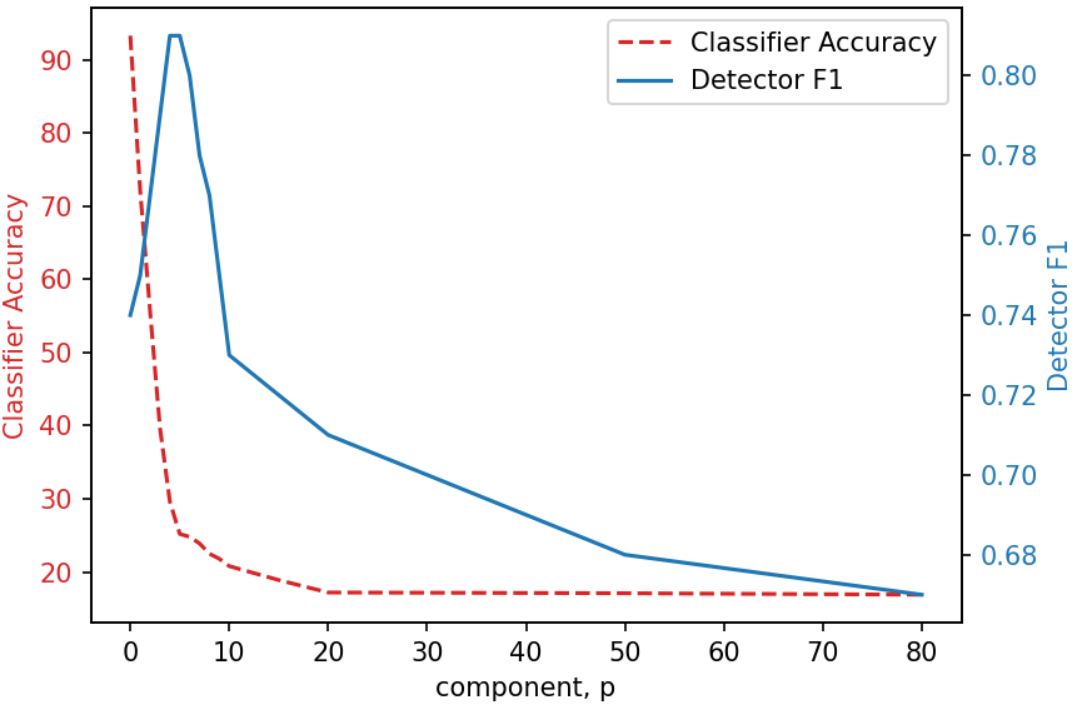}
    \caption{Performance of classifier and detector with windowed projection of encoder embedding space.}
    \label{fig:window}
\end{figure}

The aim of this section is verify that the success of the residue detector can be explained by the two main hypotheses made in Section \ref{sec:residue}. The claim that residue is left by adversarial samples in the central PCA eigenvector components is explored first. For each NLP task a PCA projection matrix is learnt in the encoder embedding space using the original training data samples (Table \ref{tab:data}). Using the test data, the residue in the embedding space can be visualized through a plot of the average (across the data) component, $\rho_i=\frac{1}{J}\sum_{j=1}^J\rho_{i,j}$ in each eigenvector direction, $\mathbf q_i$ of the original and attacked data, where
\begin{equation} \label{eqn:comp}
    \rho_{i,j} = \left |\mathcal F_{\tt en}(\mathbf x_j)^{\tt T}\mathbf q_i\right |,
\end{equation}
with $\mathbf x_j$ being the $j$th data point. Figure \ref{fig:nlp decomp} shows an example plot for the Twitter dataset, where $\rho_i$ is plotted against the eigenvalue rank, $i$ for the original and attacked data examples. Residue plots for other datasets are included in  \ref{sec:app-res-plots}. Next, it is necessary to verify that the residue detector specifically uses the residue in the central eigenvector components to distinguish between original and adversarial samples. To establish this, each encoder embedding, $\mathcal F_{\tt{en}}(x)$'s components not within a target subspace of PCA eigenvector directions $\{\mathbf q_i\}_{i=p}^{p+w}$, are removed, i.e. we have a projected embedding, $\mathbf x^{(\text{p})} = F_{\tt{en}}(x) - \sum_{i\notin [p,p+w)}\mathbf q_i^TF_{\tt{en}}(x)\mathbf q_i$,
where $w$ is a window size to choose. Now, using $\mathcal F_{\tt{cl}}(\mathbf x^{(\text{p})})$ and a residue detector trained using the modified embeddings, $\mathbf x^{(\text{p})}$, the classifier's ($\mathcal F_{\tt{cl}}(\mathbf x^{(\text{p})})$)  accuracy and detector performance (measured using F1 score) can be found. Figure \ref{fig:window} shows the performance of the classifier ($\mathcal F_{\tt{cl}}(\mathbf x^{(\text{p})})$) and the detector for different start components $p$, with the window size, $w=5$. It is clear that the principal components hold the most important information for classifier accuracy, but, as hypothesised in Section \ref{sec:residue}, it is the more central eigenvector components that hold the most information useful for the residue detector, i.e. the subspace defined by $\{\mathbf q_i\}_{i=5}^{10}$ holds the most detectable residue from adversarial examples.

The second hypothesis in Section \ref{sec:residue} claims that the existence of residue in the central eigenvector components is due to the discrete nature of NLP adversarial attacks. Hence, to analyze the impact of the discrete aspect of the attack, an artificial continuous space attack was constructed for the Twitter NLP system, where the continuous input embedding layer space (Equation \ref{eqn:embedding}) of the system is the space in which the attack is performed. Using the Twitter emotion classifier, a PGD (Equation \ref{eqn:pgd}) attack was performed on the input embeddings for each token, where the perturbation size was limited to be $\epsilon=0.1$ in the l$_\infty$ norm, achieving a fooling rate of 0.73. Note that this form of attack is artificial, as a real adversary can only modify the discrete word sequence (Equation \ref{eqn:wordsequence}). To compare the influence of discrete and continuous attacks on the same system, the average (across dataset) l$_2$ and l$_\infty$ norms of the perturbations in the input layer embedding space were found. Further, a single value summary, $N_{\sigma}$, of the residue plot (e.g. Figure \ref{fig:nlp decomp}), was calculated for each attack. $N_\sigma$ is the average difference in standard deviations between the \textit{original} component mean, $\rho_i^{\text{(orig)}}$ and \textit{attack} mean, $\rho_i^{\text{(attack)}}$,

\begin{equation}\label{eqn:std}
    N_{\sigma} = \frac{1}{I}\sum_{i=1}^I\frac{\left |\rho_i^{\text{(attack)}} - \rho_i^{\text{(orig)}}\right |}{\sqrt{ \textrm{Var}_j[\rho_{i,j}^{\text{(orig)}}]}}.
\end{equation}

\noindent Table \ref{tab:discrete vs cont} reports these metrics for the discrete and artificial continuous NLP adversarial attacks on the Twitter system~\footnote{Similar trends were found across all datasets (Table \ref{tab:disc-vs-cont-full})}. It is apparent that perturbation sizes for the discrete attacks are significantly larger. Moreover, $N_{\sigma}$ is significantly smaller for the continuous space attack, indicating that the residue left by continuous space adversarial attacks is smaller.

\begin{table}[h!]
    \centering
    \begin{small}
    \begin{tabular}{lrrr}
    \toprule
        Attack  & $N_\sigma$ & l$_2$ error & l$_\infty$ error \\ \midrule
        Discrete  &1.201& $50.2_{\pm 19.2}$ & $3.26_{\pm 0.86}$\\
        Continuous  &0.676& $5.35_{\pm 3.95}$ & $0.08_{\pm 0.03}$\\
        \bottomrule
    \end{tabular}
    
    \end{small}
    \caption{Comparison of token level \textit{discrete} attack and input embedding layer \textit{continuous} PGD attack.}
    \label{tab:discrete vs cont}
\end{table}

To explicitly observe the impact of the nature of data on detectors, adversarial attacks are considered in four domains: the discrete, sequential NLP input space (NLP-disc); the artificial continuous, sequential embedding space of an NLP model (NLP-cont); the continuous, static image input space (Img-cont) and a forced discretised, static image input space (Img-disc). For the NLP-disc and NLP-cont the same attacks as in Table \ref{tab:discrete vs cont} are used. For the continuous image domain (Img-cont), a VGG-16 architecture image classifier trained on CIFAR-100~\citep{cifar100} image data (achieving a top-5 accuracy of 90.1\%) and attacked using a standard l$_\infty$ PGD approach (Equation \ref{eqn:pgd}) is used. For the discrete image domain (Img-disc), the CIFAR-100 images, $\mathbf X \in \mathbb Z_{256}^{R\times R}$ were discretised using function $\mathcal Q_q:\mathbb Z_{256}^{R\times R}\to \mathbb Z_q^{R\times R}$, where $\mathbb Z_q = \{0, 1\frac{256}{q-1}, 2\frac{255}{q-1}, \hdots, 255\}$. In this work 2-bit quantization was used, i.e. $q=4$. With this quantization, a VGG-16 architecture was trained to achieve 78.2\% top-5 accuracy. To perform a discrete space attack, a variant of the PWWS synonym substitution attack (Section \ref{sec:attack}) was implemented, where synonyms were interpreted as closest permitted quantisation values and $N$ pixel values were substituted. For these different domains, Table \ref{tab:comparison} compares applicable  detection approaches (certain NLP detection approaches are not valid outside the word sequence space) using the best F$_1$ score, where different attack perturbation sizes are considered ($N$ substitutions for discrete attacks and for continuous attacks $|\bm\delta|\leq\epsilon$ for perturbation $\bm\delta$).

\begin{table}[htb!]
    \centering
    \begin{small}
    \begin{tabular}{l|c|rrr}
    \toprule
        Domain & Attack & Res & Unc & MD \\ \midrule
        \multirow{2}{*}{NLP-disc} & $N$=3 &\textbf{0.80} & 0.74&0.73\\
        & $N$=6 & \textbf{0.84}& 0.78&0.77\\ \midrule
        \multirow{2}{*}{NLP-cont} & $\epsilon$=0.1 & 0.67& \textbf{0.71}&0.70\\
        & $\epsilon$=0.3 & 0.67& 0.80&\textbf{0.85}\\ \midrule
        \multirow{2}{*}{Img-disc} & $N$=200 & \textbf{0.78}& 0.67&0.70\\
        & $N$=400 & \textbf{0.84}& 0.68&0.72\\ \midrule
        \multirow{2}{*}{Img-cont} & $\epsilon$=12 & 0.68& 0.70&\textbf{0.72}\\
        & $\epsilon$=48 & 0.83& 0.81&\textbf{0.87}\\
        \bottomrule
    \end{tabular}
    \end{small}
    \caption{Portability of detection approaches.}
    \label{tab:comparison}
\end{table}

In the discrete domains, the residue detection approach is better than all the other approaches. However, in the continuous data type domains, the Mahalanobis Distance dominates as the detection approach, with the residue detection approach performing the worst. As predicted by the second hypothesis of Section \ref{sec:residue}, the lack of success of the residue detection approach is expected here - the residue detection approach is only successful for discrete space attacks.

To verify that the residue detection approach is agnostic to the type of attack, the residue detector trained on \textit{substitution} attack examples was evaluated on \textit{concatenation} attack examples. Using the Twitter dataset, a $N=3$ concatenation attack was applied, achieving a fooling rate of 0.59. In this setting, the residue detector (trained on the $N=6$ substitution adversarial examples) achieved a F$_1$ score of 0.81, which is comparable to the original score of 0.84 (from Table \ref{tab:detect_perf}). This shows that even with different attack approaches similar forms of residue are produced, meaning a residue detector can be used even without knowledge of the type of adversarial attack.
\section{Conclusions}

In recent years, deep learning systems have been deployed for a large number of tasks, ranging from the image to the natural language domain. However, small, imperceptible adversarial perturbations at the input, have been found to easily fool these systems, compromising their validity in high-stakes applications. Defence strategies for deep learning systems have been extensively researched, but this research has been predominantly carried out for systems operating in the image domain. As a result, the adversarial detection strategies developed, are inherently tuned to attacks on the continuous space of images. 
This work shows that these detection strategies do not necessarily transfer well to attacks on natural language processing systems. Hence,  an adversarial attack detection approach is proposed that specifically exploits the discrete nature of perturbations for attacks on discrete sequential inputs. 

The proposed approach, termed residue detection, demonstrates that imperceptible attack perturbations on natural language inputs tend to result in large perturbations in word embedding spaces, which result in distinctive residual components. These residual components can be identified using a simple linear classifier. This residue detection approach was found to out-perform both detection approaches ported from the image domain and other state of the art NLP specific detectors. 

The key finding in this work is that the nature of the data (e.g. discrete or continuous) strongly influences the success of detection systems and hence it is important to consider the domain when designing defence strategies.

\section{Limitations, Risks and Ethics}

A limitation of the residue approach proposed in this work is that it requires training on adversarial examples, which is not necessary for other NLP detectors. This means there is a greater computational cost associated with this detector. Moreover, associated with this limitation is a small risk, where in process of generating creative adversarial examples to build a robust residue detector, the attack generation scheme may be so strong that it can more easily evade detection from other existing detectors already deployed in industry. There are no further ethical concerns related to this detector.

\section{Acknowledgements}

This paper reports on research supported by Cambridge Assessment, University of Cambridge. Thanks to Cambridge English Language Assessment for support and access to the Linguaskill-Business data. The authors would also like to thank members of the ALTA Speech Team.


\bibliography{anthology,custom}
\bibliographystyle{acl_natbib}
\newpage
\appendix
\renewcommand{\thesection}{Appendix \Alph{section}}
\section{}
\renewcommand{\thesection}{\Alph{section}}
\renewcommand\thefigure{\thesection.\arabic{figure}} 
\setcounter{figure}{0}
\renewcommand\thetable{\thesection.\arabic{table}} 
\setcounter{table}{0}

\subsection*{Training Details}

For each NLP dataset, pre-trained base (12-layer, 768-hidden dimension, 110M parameters) Transformer encoders~\footnote{\url{https://huggingface.co/transformers/pretrained_models.html}} were fine-tuned during training. Table \ref{tab:hyper-param} gives the training hyperparameters: learning rate (lr), batch size (bs) and the number of training epochs. In all training regimes an Adam optimizer was used. With respect to hardware, NVIDIA Volta GPU cores were used for training all models.

\begin{table}[h!]
    \centering
    \begin{tabular}{lrrrr}
        \toprule
        Dataset &  Model & lr & bs & epochs\\\midrule
         IMDB & BERT & 1e-5 & 8 & 2\\
         Twitter & ELECTRA & 1e-5 & 8 & 2\\
         AG News & BERT & 1e-5 & 8 & 2\\
         DBpedia & ELECTRA & 1e-5 & 8 & 2\\
         L-Bus & BERT & 1e-6 & 16 & 5\\
         \bottomrule
    \end{tabular}
    \caption{Training Hyperparameters}
    \label{tab:hyper-param}
\end{table}

\subsection*{Experiments} \label{sec:app-res-plots}

\begin{figure*}[h]
     \centering
     \begin{subfigure}[b]{0.3\textwidth}
         \centering
         \includegraphics[width=1.0\textwidth]{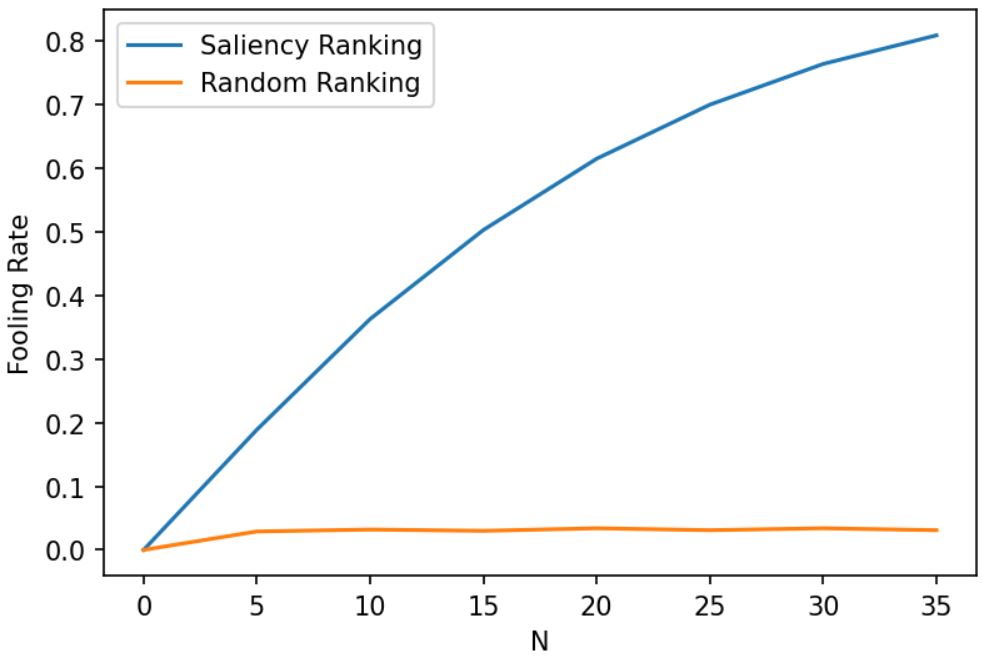}
         \caption{IMDB}
         \label{fool-IMDB}
     \end{subfigure}
     ~
     \begin{subfigure}[b]{0.3\textwidth}
         \centering
         \includegraphics[width=1.0\textwidth]{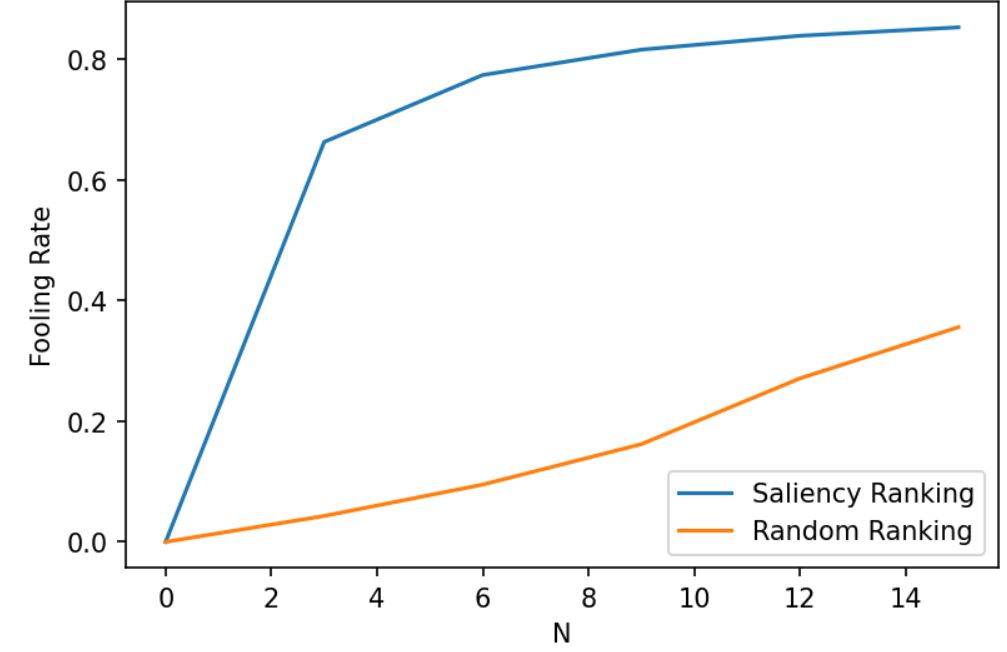}
         \caption{Twitter}
         \label{fool-Twitter}
    \end{subfigure}
    ~
     \begin{subfigure}[b]{0.3\textwidth}
         \centering
         \includegraphics[width=1.0\textwidth]{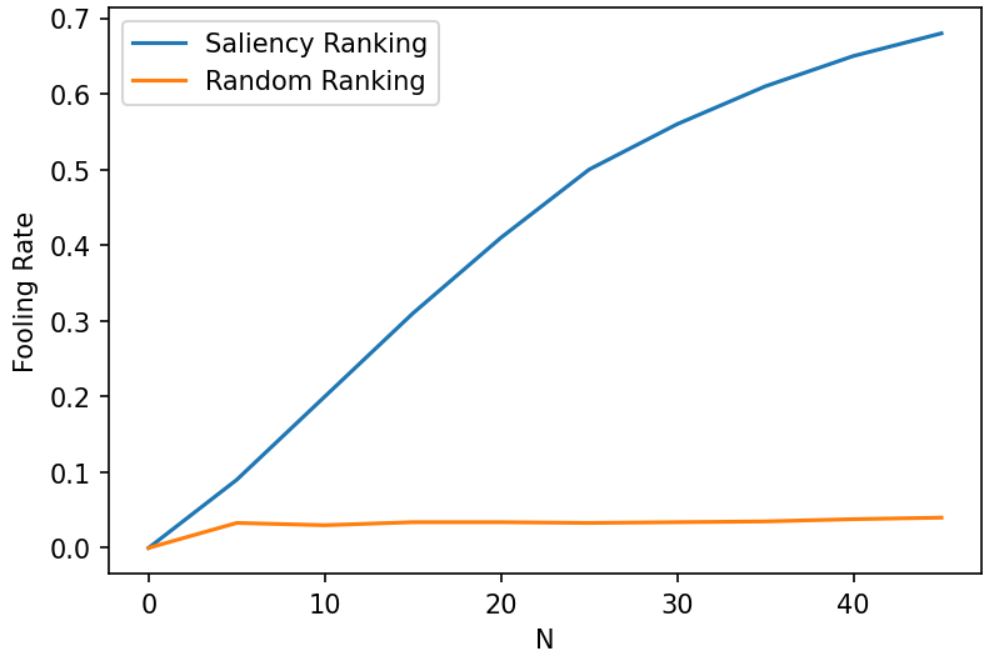}
         \caption{AG News}
         \label{fool-AG News}
    \end{subfigure}
    ~
     \begin{subfigure}[b]{0.3\textwidth}
         \centering
         \includegraphics[width=1.0\textwidth]{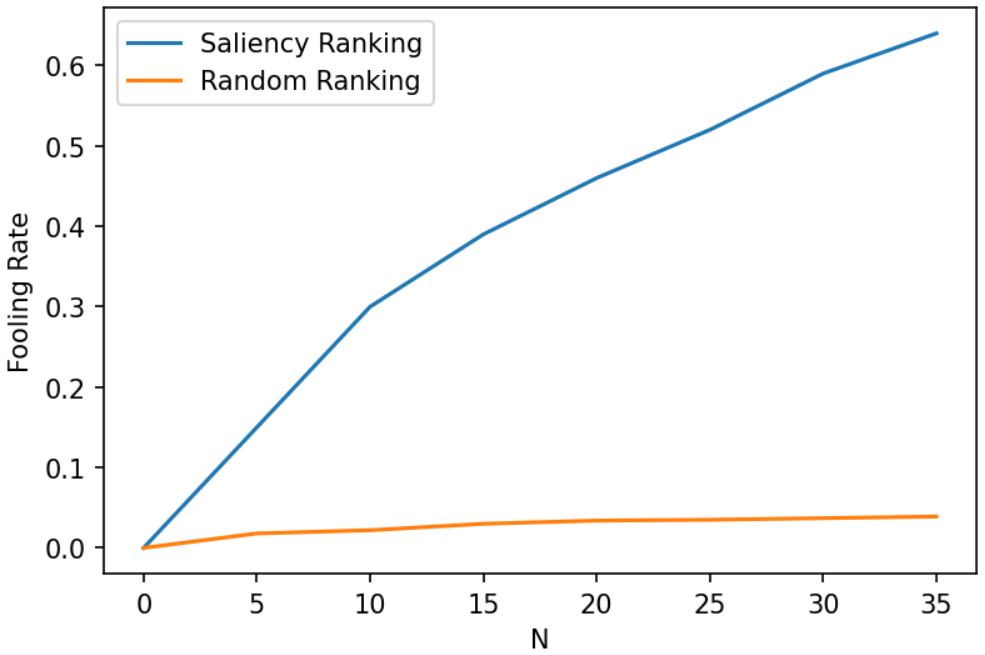}
         \caption{DBpedia}
         \label{fool-DBpedia}
    \end{subfigure}
    ~
     \begin{subfigure}[b]{0.3\textwidth}
         \centering
         \includegraphics[width=1.0\textwidth]{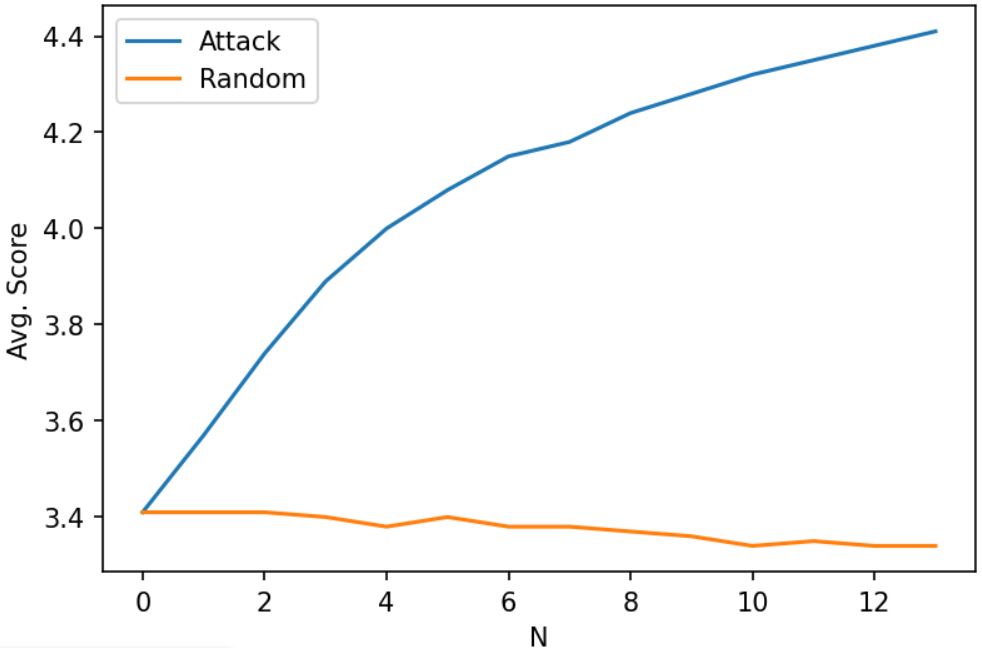}
         \caption{L-Bus}
         \label{fool-L-Business}
    \end{subfigure}
        \caption{Adversarial Attack Impact against $N$-word attack.}
        \label{fig:att}
\end{figure*}

\begin{figure*}[h] \label{fig:remaining-plots}
     \centering
     \begin{subfigure}[b]{0.3\textwidth}
         \centering
         \includegraphics[width=1.0\textwidth]{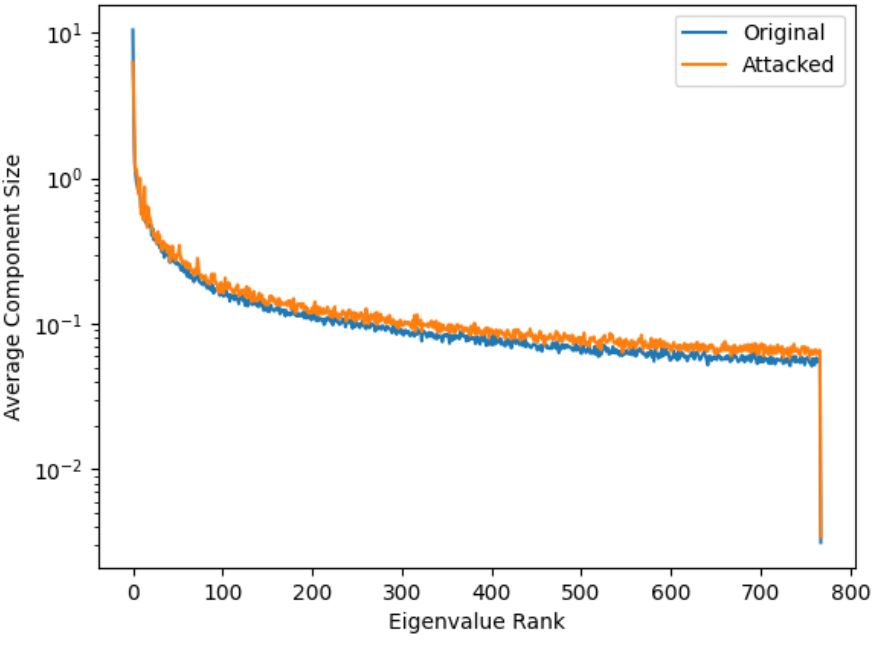}
         \caption{IMDB}
         \label{pca-IMDB}
     \end{subfigure}
     ~
     \begin{subfigure}[b]{0.3\textwidth}
         \centering
         \includegraphics[width=1.0\textwidth]{Figures/Emotion_avg_comps.JPG}
         \caption{Twitter}
         \label{pca-Twitter}
    \end{subfigure}
    ~
     \begin{subfigure}[b]{0.3\textwidth}
         \centering
         \includegraphics[width=1.0\textwidth]{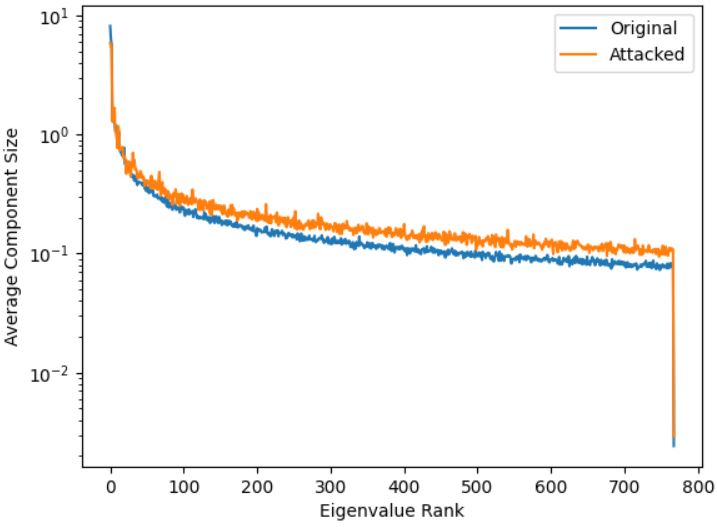}
         \caption{AG News}
         \label{pca-AG News}
    \end{subfigure}
    ~
     \begin{subfigure}[b]{0.3\textwidth}
         \centering
         \includegraphics[width=1.0\textwidth]{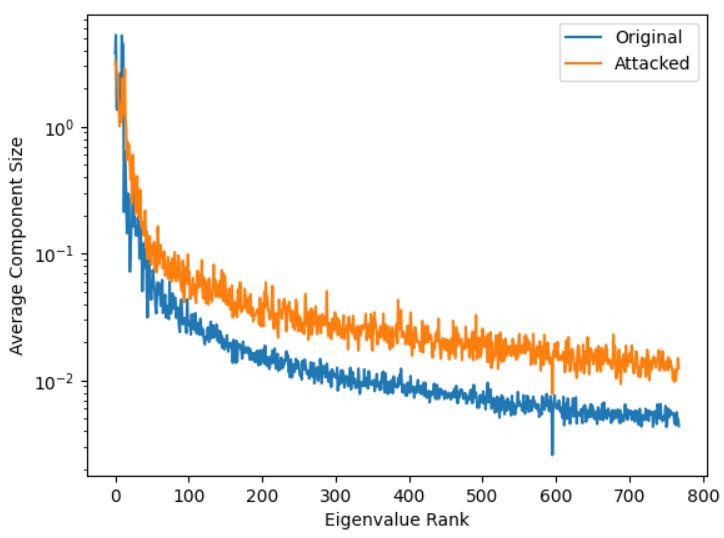}
         \caption{DBpedia}
         \label{pca-DBpedia}
    \end{subfigure}
    ~
     \begin{subfigure}[b]{0.3\textwidth}
         \centering
         \includegraphics[width=1.0\textwidth]{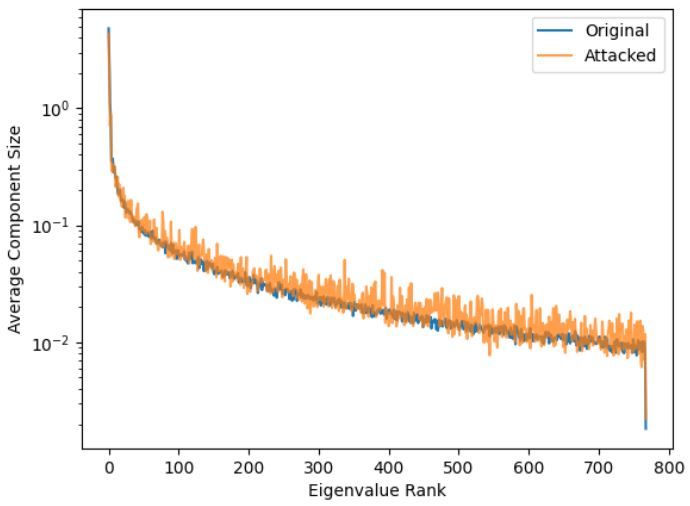}
         \caption{L-Bus}
         \label{pca-L-Business}
    \end{subfigure}
    ~
     \begin{subfigure}[b]{0.3\textwidth}
         \centering
         \includegraphics[width=1.0\textwidth]{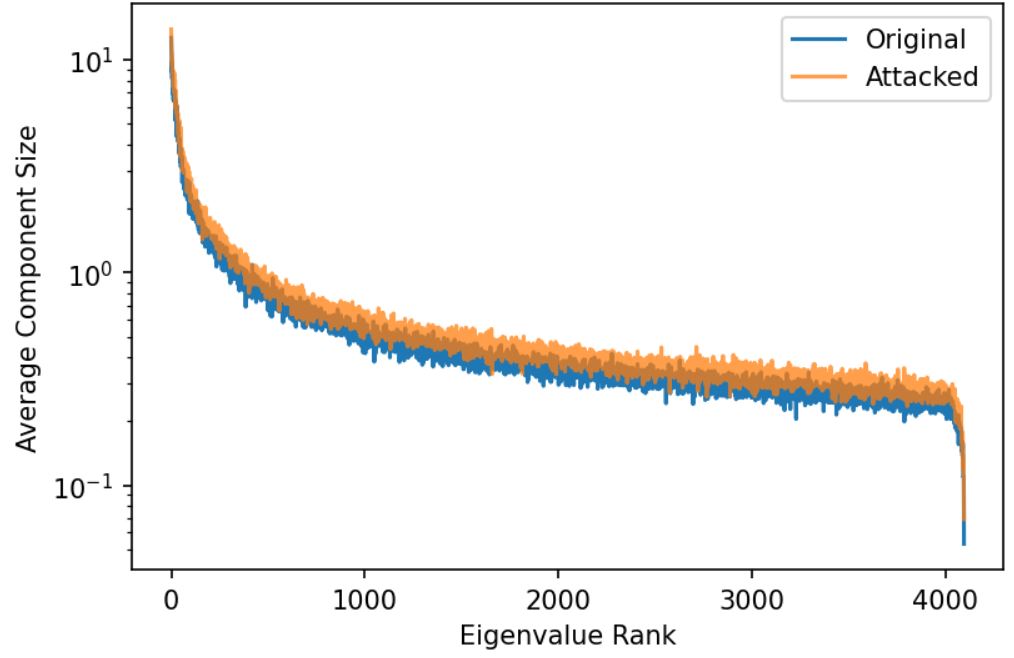}
         \caption{CIFAR-100}
         \label{pca-CIFAR-100}
    \end{subfigure}
        \caption{Encoder Embedding space residue plot using PCA decomposition.}
        \label{fig:res}
\end{figure*}

Figure \ref{fig:att} presents the impact of adversarial attacks of different perturbation sizes, $N$ on each NLP dataset. All classification datasets' models underwent saliency ranked, $N$-word substitution attacks described in Equation \ref{eqn: subst}, whilst the regression dataset, L-Bus, was subject to a $N$-word concatenation attack as in Equation \ref{eqn: concat}. For the classification tasks the impact of the adversarial attacks was measured using fooling rate, whilst for the L-Bus dataset task, the average output score from the system is given. Figure \ref{fig:res} gives the encoder embedding space PCA residue plots for all the datasets not included in the main text.




Table \ref{tab:disc-vs-cont-full} compares the impact on error sizes (using $l_2$ and $l_{\infty}$ norms) and the residue plot metric, $N_{\sigma}$ for the original text space discrete attacks and an artificial input embedding space continuous attack. The purpose of this table is to present the results for the datasets not included in the main text in Table \ref{tab:discrete vs cont}.

\begin{table*}[htb!]
    \centering
    \begin{tabular}{l|l|rrr}
    \toprule
        Dataset & Attack & $N_{\sigma}$ & $l_2$ error & $l_{\infty}$ error\\ \midrule
         \multirow{2}{*}{IMDB}& Discrete & 0.181& $73.5_{\pm 22.5}$&$4.02_{\pm 0.95}$\\
         & Continuous & 0.111& $6.73_{\pm 3.98}$& $0.09_{\pm 0.03}$\\ \midrule
         \multirow{2}{*}{Twitter}& Discrete & 1.201 & $50.2_{\pm 19.2}$ & $3.26_{\pm 0.86}$\\
         & Continuous & 0.676 & $5.35_{\pm 3.95}$ & $0.08_{\pm 0.03}$\\ \midrule
         \multirow{2}{*}{AG News}& Discrete & 0.642& $67.9_{\pm 28.1}$ & $3.35_{\pm 0.95}$\\
         & Continuous & 0.393& $5.41_{\pm 4.10}$& $0.09_{\pm 0.04}$\\ \midrule
         \multirow{2}{*}{DBpedia}& Discrete & 1.355 & $57.4_{\pm 18.6}$& $3.29_{\pm 0.88}$\\
         & Continuous & 0.991& $6.57_{\pm 3.23}$& $0.09_{\pm 0.04}$\\ \midrule
         \multirow{2}{*}{L-Bus}& Discrete & 0.201 & $94.6_{\pm 30.1}$ &$5.91_{\pm 1.12}$\\
         & Continuous & 0.135 & $8.22_{\pm 3.54}$ & $0.07_{\pm 0.04}$\\
         \bottomrule
    \end{tabular}
    \caption{Comparison of token level \textit{discrete} attack and input embedding layer \textit{continuous} PGD attack.}
    \label{tab:disc-vs-cont-full}
\end{table*}

\end{document}